\documentclass[letterpaper, 10 pt, conference]{ieeeconf}  
\IEEEoverridecommandlockouts 

\usepackage{graphicx} 

\usepackage{algorithm}
\usepackage{algpseudocode}
\usepackage{amsfonts}
\usepackage[dvipsnames]{xcolor}

\usepackage{caption}
\usepackage{amsmath}
\usepackage[backend=biber, backref=true]{biblatex}
\usepackage{hyperref}

\title{\LARGE \bf
Imagine-2-Drive: Leveraging High-Fidelity World Models via Multi-Modal Diffusion Policies
}

\author{Anant Garg$^{1}$, K. Madhava Krishna$^{1}$
\thanks{$^{1}$Robotics Research Center, IIIT Hyderabad, India. {\tt\small {garg.anant205@gmail.com, mkrishna@iiit.ac.in}}}
}

\newcommand{\coolname}{\textit{Imagine-2-Drive}}
\newcommand{\notcool}{\textit{Imagine-2-Drive }}

\makeatletter
\let\@oldmaketitle\@maketitle
\renewcommand{\@maketitle}{\@oldmaketitle
\centering
\vspace{-2mm}

\label{fig:teaser}}

\begin{document}

\maketitle

\begin{abstract}

World Model-based Reinforcement Learning (WMRL) enables sample efficient policy learning by reducing the need for online interactions which can potentially be costly and unsafe, especially for autonomous driving. However, existing world models often suffer from low prediction fidelity and compounding one-step errors, leading to policy degradation over long horizons. Additionally, traditional RL policies, often deterministic or single Gaussian-based, fail to capture the multi-modal nature of decision-making in complex driving scenarios. To address these challenges, we propose \textit{Imagine-2-Drive}, a novel WMRL framework that integrates a high-fidelity world model with a multi-modal diffusion-based policy actor. It consists of two key components: \textit{DiffDreamer}, a diffusion-based world model that generates future observations simultaneously, mitigating error accumulation, and \textit{DPA (Diffusion Policy Actor)}, a diffusion-based policy that models diverse and multi-modal trajectory distributions. By training \textit{DPA} within \textit{DiffDreamer}, our method enables robust policy learning with minimal online interactions. We evaluate our method in CARLA using standard driving benchmarks and demonstrate that it outperforms prior world model baselines, improving Route Completion and Success Rate by 15\% and 20\% respectively.

Project page: \href{https://imagine-2-drive.github.io/}{\textcolor{red}{https://imagine-2-drive.github.io/}}
\end{abstract}

\section{INTRODUCTION}\label{sec:introduction}
World models (WMs) have emerged as a powerful paradigm in reinforcement learning (RL), enabling agents to learn environment dynamics and simulate future states for improved decision-making. By leveraging a learned model of the world, RL agents can achieve greater sample efficiency, plan ahead, and generalize better across tasks. These models also enable autonomous vehicles (AVs) to internally “imagine” possible future scenarios, facilitating more efficient exploration and reducing the risks and costs associated with real-world interactions. The significant advantages of world models also highlights the importance of learning an accurate world model. 

Current WMRL \cite{hafner2020dreamcontrollearningbehaviors, hafner2022masteringataridiscreteworld, hafner2024masteringdiversedomainsworld, pan2022isodreamisolatingleveragingnoncontrollable} approaches, model the environment dynamics in latent space using a Recurrent State Space Model (RSSM) \cite{hafner2019learninglatentdynamicsplanning} based network. A common limitation of these approaches is their reliance on a single-step transition model, where errors accumulate over multi-step planning, causing planned states to drift from the on-policy distribution. To alleviate this issue, prior works \cite{wang2023drivedreamerrealworlddrivenworldmodels, ding2024diffusionworldmodelfuture, gao2024vistageneralizabledrivingworld} leverage video diffusion \cite{blattmann2023stablevideodiffusionscaling} based approaches to predict the future states simultaneously, with \textit{Vista} \cite{gao2024vistageneralizabledrivingworld} being the most versatile and accurate.
\begin{figure}[!h]
    \centering
    \includegraphics[width=1\linewidth]{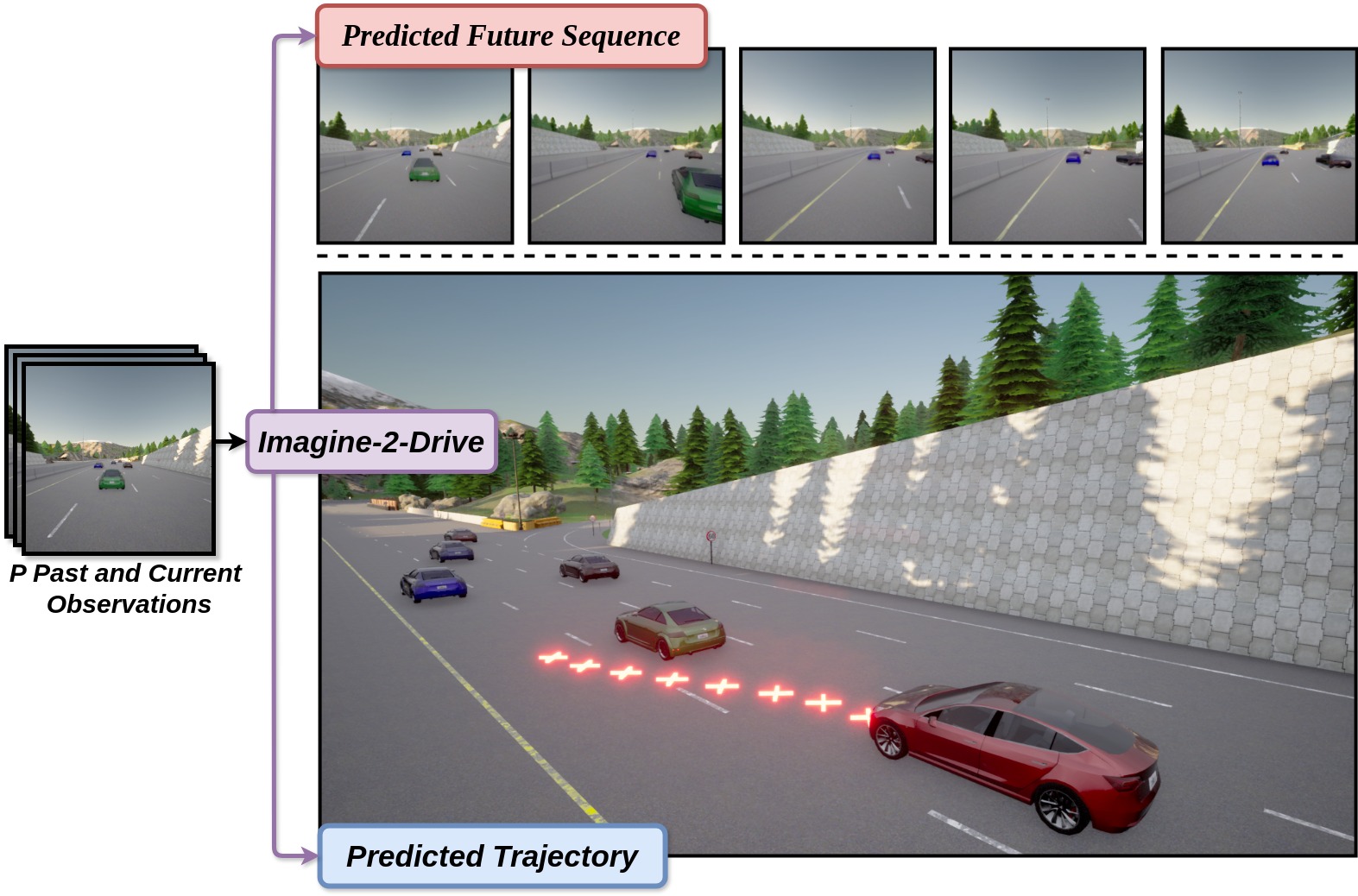}
    \caption{Based on past and current observations, \textit{Imagine-2-Drive} generates a waypoint trajectory using \textit{DPA}. This trajectory, along with observations, is used to generate future observations and rewards from \textit{DiffDreamer} corresponding to the trajectory.}
    \label{fig:teaser_figure}
\end{figure}
However, the utility of these world models is often limited by the nature of traditional RL policies. Most RL policies are constrained to deterministic outputs or single gaussian distributions, which fail to capture the full range of possible behaviors. This undermines the adaptability of the world models and their ability to handle the complexity and variability found in driving environments.

To overcome these limitations, we present \textbf{\coolname}, a novel framework that incorporates two key innovations: \textit{DiffDreamer}, a high-fidelity world model designed for precise future prediction, and \textit{DPA}, a diffusion-based policy actor that can model diverse behavioral modes for trajectory planning as shown in Fig. \ref{fig:teaser_figure}. \textit{DiffDreamer} leverages Stable Video Diffusion (SVD) architecture as a foundation for prediction of future observations with an additional module to predict future rewards which enables it to function as a comprehensive world model, facilitating more sample efficient planning and decision-making.

Similar to \cite{janner2022planningdiffusionflexiblebehavior, saha2023edmpensembleofcostsguideddiffusionmotion, jiang2023motiondiffusercontrollablemultiagentmotion, yang2024diffusionesgradientfreeplanningdiffusion}, \textit{DPA} utilizes a diffusion based model to predict the trajectory (a sequence of actions) simultaneously.
By incorporating the diverse behavior patterns inherent to diffusion policies, our framework can explore and model a broader range of behaviors, thereby improving its overall performance and robustness. \textit{DPA} is trained using PPO \cite{schulman2017proximalpolicyoptimizationalgorithms}, leveraging \textit{DiffDreamer} to simulate and evaluate trajectories. This enables optimizing the episodic return while significantly reducing the need for online interactions with the environment.


We validate our framework in the autonomous driving domain using the CARLA \cite{dosovitskiy2017carlaopenurbandriving} simulator, demonstrating its superiority over existing world model and model free approaches. A comprehensive evaluation across diverse driving scenarios, based on standard driving metrics, highlights the contributions of each component and their synergistic effect in enhancing overall performance.




To summarize, our key contributions include:
\begin{enumerate}




    \item \textbf{High-fidelity video-diffusion based world model:} We introduce a video diffusion based world model, operating solely on image inputs. Unlike traditional RSSM-based single-step world models, our approach achieves superior prediction fidelity and model accuracy, as demonstrated through comprehensive comparisons in Tables \ref{table:FIDandFVD} and \ref{table:Ablation-Driving-metrics}.


    \item \textbf{Multi-Modal Diffusion Policy Actor:} We introduce a diffusion-based policy trained that effectively captures diverse behavioral modes, overcoming the limitations of single-Gaussian or deterministic policies. As shown in Table \ref{table:Ablation-Driving-metrics} and Fig. \ref{multi-modal-result}, our approach demonstrates superior performance in modeling distinct behavioral patterns within a single policy architecture.

    \item Our extensive evaluation on CARLA driving tasks demonstrates that our framework outperforms both existing model-based approaches \cite{pan2022isodreamisolatingleveragingnoncontrollable, hafner2024masteringdiversedomainsworld} and model-free baselines \cite{mnih2013playingatarideepreinforcement, schulman2017proximalpolicyoptimizationalgorithms} across standard driving metrics, with all methods using 
 single RGB image as the sole input modality.
\end{enumerate}

\section{RELATED WORK} \label{relatedwork}
\subsection{\textbf{World Models for Autonomous Driving}}

For autonomous driving, world models follow two key paradigms: one leverages world models as neural driving simulators, while the other unifies action prediction with future generation. Think-2-Drive \cite{li2024think2driveefficientreinforcementlearning} adapts Dreamer-V3 \cite{hafner2024masteringdiversedomainsworld} for autonomous driving using BEV state-space representations.
Generative world models have been extensively explored for future driving sequence prediction based on various input modalities. Models like DriveGAN \cite{kim2021drivegancontrollablehighqualityneural}, DriveDreamer \cite{wang2023drivedreamerrealworlddrivenworldmodels}, and MagicDrive \cite{gao2024magicdrivestreetviewgeneration} focus on generating future driving scenarios using actions as inputs. GAIA-I \cite{hu2023gaia1generativeworldmodel} expands this approach by incorporating text commands alongside actions. \textit{Vista} \cite{gao2024vistageneralizabledrivingworld}, the most versatile, accepts inputs including actions, trajectories, text commands, and goal points, demonstrating high generalizability for sequence prediction.

\subsection{\textbf{Diffusion Policy for Planning}}
Following the success of Diffuser \cite{janner2022planningdiffusionflexiblebehavior} on D4RL benchmark \cite{fu2021d4rldatasetsdeepdatadriven}, diffusion-based policies \cite{chi2024diffusionpolicyvisuomotorpolicy, reuss2023goalconditionedimitationlearningusing, ankile2024juicerdataefficientimitationlearning, sridhar2023nomadgoalmaskeddiffusion, pearce2023imitatinghumanbehaviourdiffusion} have been successfully applied in robotics for motion planning.  Most typically, these policies are trained from human demonstrations through a supervised objective, and enjoy both high training stability and strong performance in modeling complex and multi-modal trajectory distributions.
In the autonomous driving domain, successful works like \cite{yang2024diffusionesgradientfreeplanningdiffusion, jiang2023motiondiffusercontrollablemultiagentmotion} sample trajectories using diffusion models while \cite{liu2024ddmlagdiffusionbaseddecisionmaking} formulate Constrained MDP (CMDP) to incorporate constraints.

\subsection{\textbf{Training Diffusion Models with Reinforcement Learning}}
Training diffusion models using reinforcement learning (RL) techniques has gained traction in recent research, particularly for applications such as text-to-image generation \cite{fan2023dpokreinforcementlearningfinetuning, black2024trainingdiffusionmodelsreinforcement, clark2024directlyfinetuningdiffusionmodels}. DDPO \cite{black2024trainingdiffusionmodelsreinforcement} formulate the denoising process as an MDP and apply PPO update to this formalism. DPPO \cite{ren2024diffusionpolicypolicyoptimization} extends these previous approaches by embedding the denoising policy MDP within the environmental MDP, forming a "dual" MDP framework optimized for various model-free tasks.

By harnessing the multi-modal capabilities of diffusion policies and the efficiency gains of world models, we investigate their combined impact and assess the individual contribution of each component to overall performance.

\begin{figure*}[!t]
\centering
\includegraphics[width=0.95\textwidth]{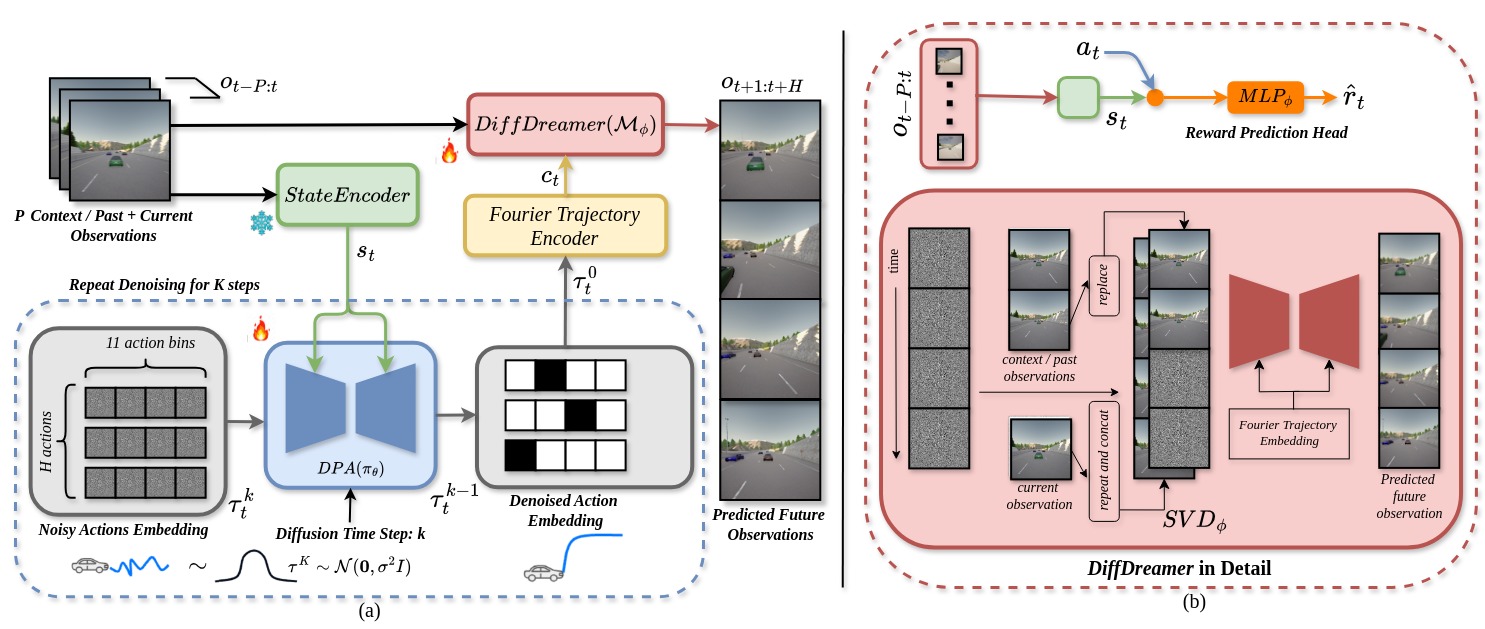}
\caption{\textbf{Architecture}: \notcool consists of a \textit{Diffusion Policy Actor (DPA)} $\pi_{\theta}$ for trajectory prediction $\tau$ and \textit{DiffDreamer} $\mathcal{M}{\phi}$ as a World Model for future state and reward prediction. (a) illustrates the overall pipeline: given the encoded state from the current and \textit{P} past observations, $\pi_{\theta}$ denoises a set of one-hot embeddings over $K$ steps to generate $H$ future discrete actions, forming the final denoised trajectory $\tau_{t}^{0}$. This trajectory is further enriched using Fourier Embeddings and, along with past and current observations, is input to $\mathcal{M}_{\phi}$ to predict future $H$ observations and rewards. (b) details \textit{DiffDreamer}, comprising two components: \textit{SVD} for future observation prediction and an additional head for reward prediction. In SVD, the first $P$ noisy frames are replaced with past observations, while the current observation is repeated $(P+H)$ times and concatenated with past and noisy frames for better grounding with the initial conditions.}
\label{fig:architecture_pipeline}

\end{figure*}

\section{METHODOLOGY} \label{methodology}



\notcool is a world model-based RL framework for long-horizon trajectory generation using only front-facing camera input. By leveraging a learned world model for policy learning, it optimizes for maximizing the episodic returns while minimizing reliance on online environment interactions, enabling efficient decision-making.


As depicted in Fig. \ref{fig:architecture_pipeline}(a), the framework consists of three key components:
\begin{enumerate}
    \item \textbf{State Encoder:} Encodes the sequence of $P$ past front-view RGB camera observations with the current observation, to provide temporal information for the current state $s$.

    \item \textbf{\textit{DPA}:} A diffusion-based policy actor which generates a sequence of actions (waypoint trajectory $\tau$) conditioned on the current encoded state. 

    \item \textbf{\textit{DiffDreamer}:} A high-fidelity world model that improves sample efficiency by simulating future states, rewards, and synthetic experiences. It generates future observations based on past frames and the \textit{DPA}-predicted waypoint trajectory, offering a rich predictive model for policy learning.

\end{enumerate}

\textbf{Notation:} This paper differentiates between two categories
of timesteps: diffusion timesteps, indicated by superscripts
$k \in \{K, \dots, 0\}$, and environment timesteps, denoted by subscripts $t \in \{1, \dots, T\}$. We use ( $\hat{}$ ) above elements to denote them being predicted in the future.

We also use the terms world model and \textit{DiffDreamer} interchangeably, as they refer to the same concept over here.

\subsection{\textbf{Approximate POMDP with MDP}}
A single-frame state representation in autonomous driving lacks temporal context, limiting its formulation as an MDP, which assumes the state fully captures relevant information. To address this, we approximate an MDP by incorporating a fixed-length sequence of past observations, enriching temporal dependencies and mitigating partial observability.

Following ViNT \cite{shah2023vintfoundationmodelvisual}, we use a pretrained \textit{State Encoder} which tokenize each observation image $\{o_{i}\}_{t-P}^{t}$ into an embedding of size $d_{model} = 512$ with an EfficientNet-B0 \cite{tan2020efficientnetrethinkingmodelscaling} model which outputs a feature vector $\zeta(o_{i})$. The individual tokens are then combined with a positional encoding and fed into a transformer backbone $\mathcal{F}_{sa}$. We use a decoder-only transformer with $n_{L}=4$ multi-headed attention blocks, each with $n_{H}=4$ heads and $d_{FF}=2048$ hidden units. The output tokens are concatenated and flattened, then passed through MLP layers to give a final state embedding $s_{t} \in \mathbb{R}^{32}$
\begin{equation}
    s_{t} = MLP \left( \mathcal{F}_{sa} \left( \zeta(o)_{t-P:t} \right) \right) \label{state-encoder}
\end{equation}

\subsection{\textbf{DiffDreamer}}

To overcome the issues of compounding errors and low prediction fidelity, typically found in one-step prediction RSSM-based world models, we leverage stable video-diffusion (SVD) as the foundation for building our world model $\mathcal{M}_{\phi}$ to predict multiple future observations at once. Given the $P$ past and current observations $\{o_{i}\}_{t-P}^{t}$, we predict $H$ future observations $\{\hat{o}_{i}\}_{t+1}^{t+H}$ conditioned on a sequence of actions (trajectory of waypoints) $\tau \in \mathbb{R}^{H \times 2}$ from \textit{DPA}.
\begin{equation}
    \tau_{t} = (a_{t}, a_{t+1}, \dots, a_{t+H}) \label{trajectory_equation}
\end{equation}
where action $a_{t}$ is defined in eq. \ref{action_equation} \newline


\textbf{Future State Prediction:} As shown in the Architecture Fig. \ref{fig:architecture_pipeline}(b), to predict the future observations $\{\hat{o}_{i}\}_{t+1}^{t+H}$ corresponding to $\tau_{t}$ while enforcing dynamic consistency with respect to position, velocity and acceleration, we replace the first $P$ noisy frames of the SVD model with the past $P$ and current observations $\{o_{i}\}_{t-P}^{t}$. This approach enforces the grounding of model's prediction of the initial frames with the initial conditions and reduces drift.
\begin{equation}
    \hat{o}_{t+1:t+H} = SVD_{\phi}(o_{t-P:t}, c_{t})
\end{equation}
where $c_{t}$ is the fourier embedding of trajectory $\tau_{t}$ using eq. \ref{fourier_eq}

Using the state encoder eq. \ref{state-encoder} and $\{\hat{o}_{i}\}_{t+1}^{t+H}$, we can get the corresponding future states $\{\hat{s}_{i}\}_{t+1}^{t+H}$. This capability allows the model to accurately predict the future states of the environment corresponding to the actions, which is crucial for sample efficient and safe decision making. \newline

\textbf{Fourier Trajectory Encoding:} To effectively encode the trajectory $\tau_{t}$, we apply a fourier embedding (FFT) \cite{mildenhall2020nerfrepresentingscenesneural}, which transforms the raw spatial coordinates into a frequency domain representation. This embedding captures both low and high-frequency components, allowing the model to represent smooth and periodic motion patterns more effectively.
\begin{equation}
    c_{t} = FFT(\tau_{t}) \label{fourier_eq}
\end{equation}

\textbf{Additional Required MDP Components:} To employ \textit{DiffDreamer} as a world model, we integrate an additional $MLP$ module to predict future rewards $\hat{r}_{t}$ based on the state $s_{t}$ and action $a_{t}$, thereby completing the Markov Decision Process (MDP) formulation, enabling effective policy optimization in the ``imagination" space of the world model.
\begin{equation}
\hat{r}_{t} \sim p_{\phi}\left( \hat{r}_{t} \mid \hat{s}_{t}, a_{t} \right) = MLP(\hat{s}_{t}, a_{t})
\label{eq:reward-gamma-predict}
\end{equation}

\textbf{Loss Function:} For future frame predictions, in addition to usual noise reconstruction loss $\mathcal{L}_{diffusion}$, we incorporate two additional losses $\mathcal{L}_{dynamics}$ and $\mathcal{L}_{structure}$ from section 3.1 of \cite{gao2024vistageneralizabledrivingworld} for enhanced frame prediction. A brief explanation and intuition behind the additional losses follows:
\begin{enumerate}
    \item \textbf{\textit{Dynamics Enhancement Loss}} $\mathcal{L}_{dynamics}$ focuses on accurately capturing the motion dynamics of objects within the generated videos. This loss encourages the model to learn realistic movement patterns, ensuring that the generated sequences maintain coherence over time.
    \item \textbf{\textit{Structure Preservation Loss}} $\mathcal{L}_{structure}$ reinforces the structural integrity of the generated scenes. By emphasizing the preservation of spatial relationships and object configurations, this loss helps in producing videos that are not only visually appealing but also contextually accurate.
\end{enumerate}
For a detailed overview of the losses used in $\mathcal{L}_{SVD}$, interested readers are encouraged to refer to section 3.1 of \cite{gao2024vistageneralizabledrivingworld}.

The combined loss function for training the SVD model $\mathcal{L}_{SVD}$ becomes:
\begin{equation}
    \mathcal{L}_{SVD} = \mathcal{L}_{diffusion} + \lambda_{1} \mathcal{L}_{dynamics} + \lambda_{2} \mathcal{L}_{structure} \label{SVD_loss_function}
\end{equation}

To train the additional reward prediction head, we add an additional reward prediction loss $\mathcal{L}_{r}$ to get the final world model loss $\mathcal{L}_{\mathcal{M}}$:
\begin{gather}
    \mathcal{L}_{r} = \frac{1}{2}|| \hat{r}_{t} - r_{t} ||^{2}\\
    \mathcal{L}_{\mathcal{M}} = \mathcal{L}_{SVD} + \lambda_{3} \mathcal{L}_{r} \label{world_model_loss}
\end{gather}

Here $\lambda_{1}$, $\lambda_{2}$, $\lambda_{3}$ are hyperparameters to balance the weights of the losses in the final world model loss function $\mathcal{L}_{WM}$. \newline

\subsection{\textbf{Diffusion Policy Actor (DPA)}}
We use a diffusion based policy network $\pi_{\theta}$ to predict a trajectory $\tau_{t}$ as shown in eq. \ref{trajectory_equation} conditioned on the current state $s_{t}$. $\pi_{\theta}$ employs a U-Net architecture similar to that proposed in \cite{janner2022planningdiffusionflexiblebehavior}. For sampling, we adopt the strategy from \textit{DDIM} \cite{song2022denoisingdiffusionimplicitmodels} and utilize $K=50$ denoising steps.

However, unlike behavioral cloning methods where diffusion policies can be optimized to fit the conditional noise prediction $\varepsilon_{\theta}(\tau_{t}^{k}, s_{t}, k)$, our RL setting lacks predefined target trajectories to supervise upon. Instead, the policy learns to maximize the cumulative sum of rewards objective $\mathcal{J}_{RL}(\pi_{\theta})$.
\begin{equation}
    \mathcal{J}_{RL}(\pi_{\theta}) = \mathbb{E}_{\pi_{\theta}} \biggl[\sum_{t \geq 0} r(s_{t}, a_{t}) \biggr]  \label{rl_objective}  
\end{equation}
where $r_{t}$ is calculated from the simulator using eq. \ref{reward_function_equation}.

As shown in \cite{black2024trainingdiffusionmodelsreinforcement} and \cite{fan2023dpokreinforcementlearningfinetuning}, the denoising process can be formulated as a $K$-step MDP, to form a denoising MDP. Our approach extends this approach by augmenting the environment MDP with the diffusion MDP. The denoising MDP is completed for each time-step of environment MDP to get the final denoised trajectory $\tau_{t}^{0}$. Unlike DPPO \cite{ren2024diffusionpolicypolicyoptimization}, which applies a similar formulation in a model-free setting, our approach leverages this structure within model-based reinforcement learning paradigm.

We begin with a randomly initialized diffusion policy network. Sampling from the policy network begins with a noisy trajectory sample drawn from a isotropic Gaussian distribution, $\tau^{K}_{t} \sim \mathcal{N(\textbf{0, $\sigma^{2}$I})}$. The reverse process is defined by a learned distribution $p_{\theta}(\tau^{k-1}_{t} | \tau^{k}_{t}, s_{t})$, which progressively ``denoises" the action sequence to produce a sequence of trajectories $\{ \tau^{K}_{t}, \tau^{K-1}_{t}, \dots, \tau^{1}_{t}, \tau^{0}_{t} \}$ ending with the final denoised sample $\tau^{0}_{t}$. \newline

\textbf{Denoising as a $K$-step MDP:} The Denoising MDP uses timestep $\bar{t}(t,k) = tK + (K - k)$ corresponding to (\textit{t,k}). The states, actions and rewards are defined as:
\begin{subequations}
    \begin{gather}
        \bar{s}_{\bar{t}(t,k)} = (\tau_{t}^{k+1}, s_{t}) \\
        \bar{a}_{\bar{t}(t,k)} = \tau_{t}^{k} \\
        \bar{r}_{\bar{t}(t,k)} = \begin{cases} 
        r(s_{t}, a_{t}) & \text{if } k = 0 \\
        0 & \text{otherwise}
        \end{cases}
    \end{gather}
\end{subequations}
We assign rewards only to the final denoising step, corresponding to actions $a_{t}^{0}$ executed after the completion of the denoising MDP. The policy $\pi_{\theta}$ parameterizes as:
\begin{equation}
\begin{aligned}
    \bar{\pi}_{\theta}(\bar{a}_{\bar{t}(t,k)} \mid \bar{s}_{\bar{t}(t,k)}) = \pi_{\theta}(\tau_{t}^{k} \mid \tau_{t}^{k+1}, s_{t}) \\ = \mathcal{N}(\tau_{t}^{k}, \mu(\tau_{t}^{k}, \varepsilon_{\theta}(\tau_{t}^{k+1}, k+1, s_{t}), \sigma^{2}I)) \label{pi_beta}
\end{aligned}
\end{equation}

\textbf{Policy Optimization:} 
\textit{DPA} policy parameters $\theta$ can be optimized by maximizing policy objective $\mathcal{J}_{RL}$ eq. \ref{rl_objective}, using policy gradients method:

\begin{equation}
\begin{aligned}
    \nabla_{\theta}\mathcal{J}_{RL} = \mathbb{E}_{\pi_{\theta}} \biggl[ \sum_{t \geq 0} \nabla_{\theta} log \pi_{\theta}(a_{t} \mid s_{t}) R_{t} \biggr], \\
    R_{t} = \sum_{t' \geq t}\gamma(t')r_{t'} \quad \text{is the go-to-rewards}
\end{aligned}
\end{equation}

This objective can be applied to optimize $\bar{\pi}_{\theta}$ for which the gradients of the log-likelihood can be calculated from eq. \ref{pi_beta}:
\begin{equation}
\begin{aligned}
    \nabla_{\theta}\mathcal{\bar{J}}_{RL} = \mathbb{E}_{\bar{\pi}_{\theta}} \biggl[ \sum_{\bar{t} \geq 0} \nabla_{\theta} log \bar{\pi}_{\theta}(\bar{a}_{\bar{t}} \mid \bar{s}_{\bar{t}}) \bar{R}_{\bar{t}} \biggr], \bar{R}_{\bar{t}} = \sum_{t' \geq \bar{t}}\gamma(t')\bar{r}_{t'}
\end{aligned}
\end{equation}

For better stability and reduced variance, we use PPO \cite{schulman2017trustregionpolicyoptimization} to get the final gradient estimator for $\bar{\pi}_{\theta}$ as follows:
\begin{equation}
\begin{aligned}
    \nabla_{\theta} \mathcal{\bar{J}}_{RL} &= \nabla_{\theta} \mathbb{E} \Bigg[ 
    \min \Bigg( A^{\bar{\pi}_{\theta_{\text{old}}}}(\bar{s}_{\bar{t}}, \bar{a}_{\bar{t}}) 
    \frac{\bar{\pi}_{\theta}(\bar{a}_{\bar{t}} \mid \bar{s}_{\bar{t}})}
    {\bar{\pi}_{\theta_{\text{old}}}(\bar{a}_{\bar{t}} \mid \bar{s}_{\bar{t}})}, \\ 
    &\quad A^{\bar{\pi}_{\theta_{\text{old}}}}(\bar{s}_{\bar{t}}, \bar{a}_{\bar{t}}) 
    \operatorname{clip} \biggl(\frac{\bar{\pi}_{\theta}(\bar{a}_{\bar{t}} \mid \bar{s}_{\bar{t}})}
    {\bar{\pi}_{\theta_{\text{old}}}(\bar{a}_{\bar{t}} \mid \bar{s}_{\bar{t}})}, 1-\varepsilon, 1+\varepsilon \biggr)
    \Bigg) 
    \Bigg] \label{dpa_update}
\end{aligned}
\end{equation}
where $\varepsilon$ is the clipping ratio which prevents drastic updates to the policy, leading to more stable training compared to vanilla PG.

The advantage estimator $A^{\bar{\pi}_{\theta_{\text{old}}}}(\bar{s}_{\bar{t}}, \bar{a}_{\bar{t}})$ is used to reduce the variance and stabilize training without introducing any bias. It is calculated as:
\begin{equation}
    A^{\bar{\pi}_{\theta_{\text{old}}}}(\bar{s}_{\bar{t}}, \bar{a}_{\bar{t}}) = \sum_{t' \geq t}\bar{r}(s_{\bar{t}(t,0)}, a_{\bar{t}(t,0)}) - V_{\theta}(s_{\bar{t}(t,0)}) \label{advantage_estimator}
\end{equation}

where $V_{\theta}$ is the state-value estimator, which is trained using $TD(0)$ method as follows:
\begin{equation}
    \mathcal{L}_{V_{\theta}} = \frac{1}{2}||\delta_{t}||^{2}, \quad \text{where } \delta_{t} = r + V_{\theta'}(s_{t+1}) - V_{\theta}(s_{t}) \\
\end{equation}
where $\delta_{t}$ is the $TD$ residual error.

\begin{figure}[!h]
    \centering
    \includegraphics[width=1.0\linewidth]{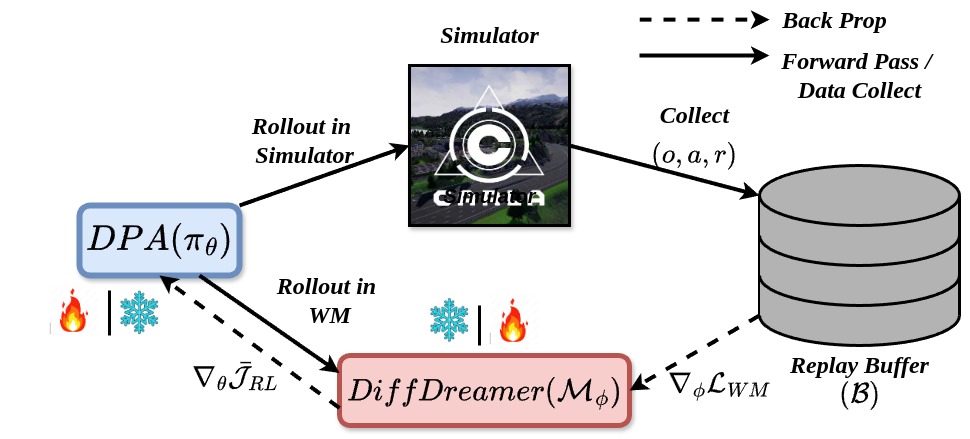}
    \caption{\textbf{Iterative Training of DPA ($\pi_{\theta}$) and DiffDreamer ($\mathcal{M}_{\phi}$):} $\pi_{\theta}$ and $\mathcal{M}_{\phi}$ are trained alternately, with one fixed while the other updates. $\mathcal{M}_{\phi}$ learns from $\pi_{\theta}$ rollouts in the simulator, while $\pi_{\theta}$ is optimized using a frozen $\mathcal{M}_{\phi}$. This iterative process ensures synchronization and improves training stability.}
    \label{fig:iterative_training}
\end{figure}

The training procedure for the overall pipeline is outlined in Algorithm \ref{alg:policy_world_model_training} and shown in Fig. \ref{fig:iterative_training}.

\begin{algorithm}
\caption{Training of \coolname}
\label{alg:policy_world_model_training}
\begin{algorithmic}[1]
\State \textbf{Initialize:} Randomly initialized policy $\pi_\theta$ and world model $\mathcal{M}_\phi$
\State Initialize experience buffer $\mathcal{B} \leftarrow \emptyset$
\State Define total online interactions $N_{\text{total\_online}} \gets 10^6$
\State Define rollout limit per env. step $N_{\text{rollout\_online}} \gets 5000$
\State Define world model warm-start steps $N_\text{world\_ws} \gets 10000$
\State Define world model update steps $N_{\text{world}} \gets 2$
\State Define policy update steps $N_{\text{policy}} \gets 5$

\For{$n = 1$ to $N_{\text{total\_online}}$}
    \State \textbf{(1) Collect Experience:}
    \For{$t = 1$ to $N_{\text{rollout\_online}}$}
        \State Sample trajectory $\tau_t \sim \pi_\theta(s_t)$
        \State Execute $a_t$ in simulator, observe $(s_t, a_t, r_{t})$
        \State Store transition $(s_t, a_t, r_{t})$ in buffer $\mathcal{B}$
        \State $n \gets n+1$
    \EndFor
    
    \State \textbf{(2) Train World Model:}
    \For{$i = 1$ to $N_{\text{world}}$}
        \State Sample batch of $\left( \{o_{i}\}_{t-P}^{t+H}, \{ a_{j}\}_{t}^{t+H}, \{r_{j} \}_{t}^{t+H} \right)$ from $\mathcal{B}$
        \State Update world model $\mathcal{M}_\phi$ using loss $\mathcal{L}_{\mathcal{M}}$ (eq. \ref{world_model_loss})
    \EndFor
    
    \State \textbf{(3) Train Policy $\pi_{\theta}$ in the 
    World Model $\mathcal{M}_{\phi}$}:
    \If{n $\geq$ $N_{world\_ws}$}
    \For{$j = 1$ to $N_{\text{policy}}$}
        \State Sample initial state $s_0$ from $\mathcal{B}$
        \State Roll out trajectory using $\mathcal{M}_\phi$:
        \For{$t = 0$ to $T$}
            \State Sample trajectory $\tau_t \sim \pi_\theta(s_t)$
            \State Predict $s_{t+1}, r_{t+1} \sim \mathcal{M}_\phi(s_t, \tau_t)$
        \EndFor
        \State Optimize policy $\pi_\theta$ using $\nabla_{\theta}\mathcal{\bar{J}}_{RL}$ (eq. \ref{dpa_update})
    \EndFor
    \EndIf
\EndFor
\end{algorithmic}
\end{algorithm} 

\begin{figure*}[!t]
    \centering
    \includegraphics[width=0.9\textwidth]{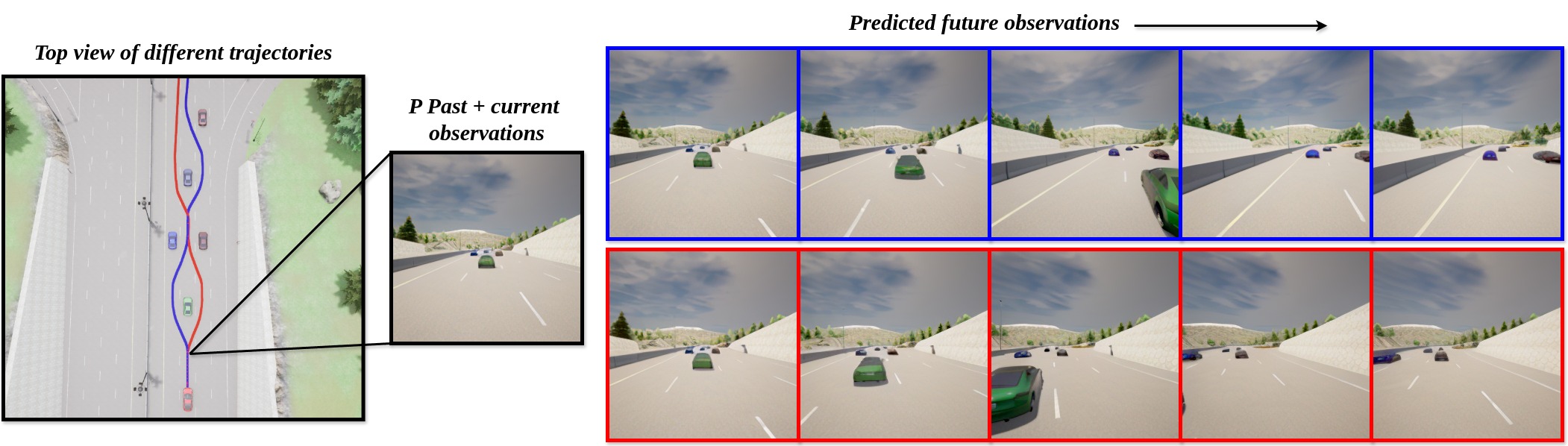}
    \caption{\textbf{Multi-Modal Nature of \textit{DPA}:} The top-view visualization highlights the diverse behaviors generated over the episode by \textit{DPA} under identical initial conditions but different seeds. Given the current and past observations as shown, \textit{DiffDreamer} predicts future observations for two distinct trajectories: \textcolor{blue}{\textbf{Left (Blue)}} and \textcolor{red}{\textbf{Right (Red)}}, navigating around the \textcolor{ForestGreen}{\textbf{Green car}} ahead. The predicted frames are color coded corresponding to the trajectory. This demonstrates both the multi-modal nature of \textit{DPA} and the prediction-fidelity of \textit{DiffDreamer}. (Please zoom-in for a better view)}
    \label{multi-modal-result}
    
\end{figure*}

\begin{figure}[!h]
    \centering
    \includegraphics[width=1.0\linewidth]{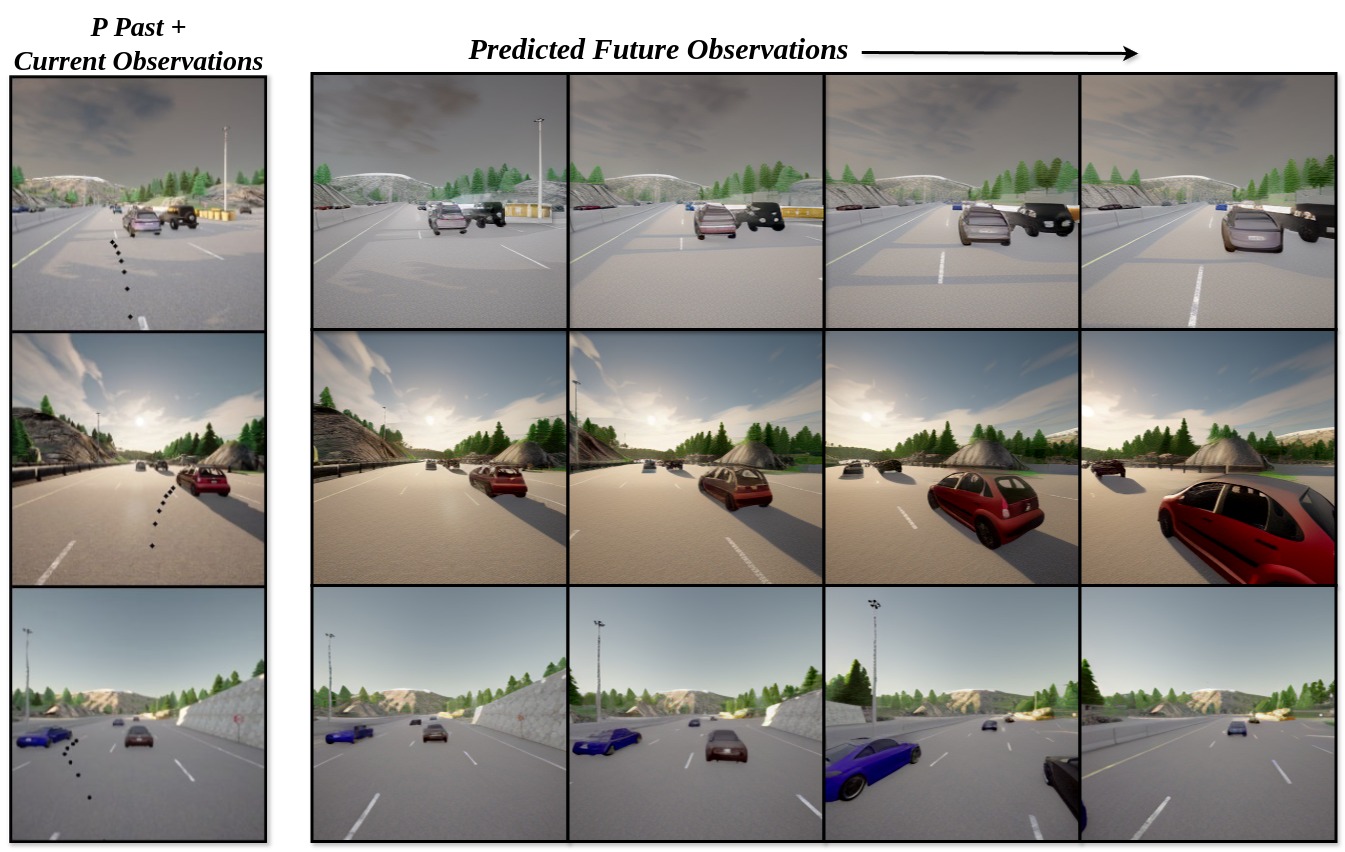}
    \caption{\textbf{\textit{DiffDreamer} Future Prediction:} Future observation predictions from \textit{DiffDreamer}, conditioned on the input trajectory (shown in \textbf{black dots $\cdots$}) and current observations. Demonstrates our world model's ability to accurately predict future observations based on the provided context, highlighting its robust trajectory prediction capabilities. (Please zoom-in for a better view)}
    \label{fig:action-controllability}
\end{figure}

\section{EXPERIMENTS AND RESULTS} \label{experiments}

\subsection{\textbf{Experimental Setup}}
We evaluate navigation along predefined routes in CARLA Town04 (v0.9.11) at 20 Hz, utilizing routes from the CARLA Leaderboard. Each scenario varies in environmental and lighting conditions, featuring up to 500 meters of road with multiple lanes and a maximum of 20 dynamic traffic vehicles. The ego-agent is randomly initialized along the route, while traffic vehicles are assigned velocities between 1–5 km/h. An episode terminates upon collision or going out-of-bounds. The ego-agent maintains a constant velocity of 7 km/h and follows the output waypoint trajectory via a default PID controller. We train and evaluate our model and baselines on a subset of 10 scenarios.


\subsection{\textbf{Training Details}}
\textit{Vista} \cite{gao2024vistageneralizabledrivingworld}, trained on approximately 1,700 hours of driving videos, serves as the foundation for initializing our world model, \textit{DiffDreamer}. To facilitate the efficient convergence of \textit{DPA}, $\mathcal{M}_{\phi}$ is provided with a warm-start by pretraining it for 10,000 iterations, which are obtained from rollouts in the CARLA simulator using a randomly initialized $\pi_{\theta}$. As the training of $\mathcal{M}_{\phi}$ relies on data tuples collected by $\pi_{\theta}$ rollouts in the simulator, it is crucial for both the modules to learn in a synchronized manner to ensure stable and efficient co-evolution of their capabilities. We train \textit{DiffDreamer} on 4 A100 GPUs for 28 hours, accumulating gradients over 4 steps for an effective batch size of 16.

\subsubsection{\textbf{Training of \textit{DiffDreamer}}}
$\mathcal{M}_{\phi}$ is trained in a supervised manner using the data tuples, where each tuple consists of $\left( \{o_{i}\}_{t-P}^{t+H}, \{ a_{j}\}_{t}^{t+H}, \{r_{j} \}_{t}^{t+H} \right)$ sampled from the replay buffer. These tuples are used to compute the world model loss $\mathcal{L}_{\mathcal{M}}$ using eq. \ref{world_model_loss}, which then updates the world model parameters $\phi$. Note that the policy $\pi_{\theta}$ is kept fixed during the training of $\mathcal{M}_{\phi}$.

\subsubsection{\textbf{Training of DPA}} The policy $\pi_{\theta}$ is trained within the ``imagination'' space of the frozen world model $\mathcal{M}_{\phi}$. Given a state $s_t$, $\pi_{\theta}$ generates a trajectory $\tau_t$. The past observations $\{o_i\}_{t-P}^{t}$ along with $\tau_t$ are then input to $\mathcal{M}_{\phi}$ to predict future observations and rewards, denoted as $(\hat{o}_i, \hat{r}_i)$ for $i = t, \dots, t+H$. While predictions extend up to an $H$-step horizon, the agent advances one timestep at a time in the world model, with the predicted reward $\hat{r}_t$ contributing to the go-to-return $R_t$. The next state $\hat{s}_{t+1}$ is then used by $\pi_{\theta}$ to generate $\tau_{t+1}$. This process iterates until a fixed episode length $T$, after which $\pi_{\theta}$ is updated using eq.
 \ref{dpa_update}.

This iterative training procedure is illustrated in Fig. \ref{fig:iterative_training}, where the policy actor $\pi_{\theta}$ interacts with the world model $\mathcal{M}_{\phi}$ in a closed-loop fashion. By leveraging the simulated rollouts generated by $\mathcal{M}_{\phi}$, $\pi_{\theta}$ refines its policy without requiring direct interaction with the simulator, thereby enhancing sample efficiency.


\subsection{\textbf{Implementation Details}}
\subsubsection{\textbf{Action Space}}
Waypoint trajectory is represented in the X-Y cartesian space. To predict the trajectory effectively, we simplify the action space by predicting the incremental $\Delta X$ and $\Delta Y$ in the ego-agent's local frame.
\begin{equation}
\begin{aligned}
    a_{t} = (\Delta X_{t}, \Delta Y_{t}) \label{action_equation}
\end{aligned}
\end{equation}
\textit{Simplifying $\Delta X$:} To reduce action space dimensionality and ensure forward progress, we fix $\Delta X = 1$, allowing the model to focus on lateral adjustments.

\textit{Discretizing $\Delta Y$:} lateral movement $\Delta Y$ per timestep is defined within $(-0.5, 0.5)$ meters, divided into 11 equal bins representing specific lateral deviations.

\textit{Action Encoding:} Each discrete action, representing $\Delta Y$, is encoded using a one-hot embedding of 11 size.
\label{Action-Space}

\subsubsection{Horizon and Context Length}
We set prediction horizon length $H=9$ and context length $P=5$.

\subsubsection{\textbf{Reward Function}}
The reward function for lane-keeping and collision avoidance is defined as:
\begin{equation}
\begin{aligned}
    r_{t} = \upsilon_{ego}^T \hat{u}_{h} \cdot \Delta t - \xi_{1} \cdot \big| Collision\_Cost \big| - \xi_{2} \cdot \big| \Delta Y \big| + c \label{reward_function_equation}
\end{aligned}
\end{equation}
Here, $\upsilon_{ego}$ represents the ego agent's velocity projected onto the highway direction $\hat{u}{h}$, normalized and scaled by $\Delta t = 0.05$ to measure highway progression in meters. Collisions with other vehicles or the environment incur a penalty of $Collision_cost = 10$, while $\xi{2} \cdot \big| \Delta Y \big|$ discourages large trajectory deviations. The constant $c=1$ promotes longer episodes, with $\xi_{1}$ and $\xi_{2}$ both set to 1.
\label{Reward-Function}
\vspace{-3px}
\subsection{\textbf{Baselines}}
We focus on world model-based and model-free methods using front camera images as the sole input modality and compare with the following:
\begin{enumerate}
    \item Dreamer-V3 \cite{hafner2024masteringdiversedomainsworld}: The third generation in the Dreamer series \cite{hafner2020dreamcontrollearningbehaviors, hafner2022masteringataridiscreteworld}, incorporating robustness and normalization techniques for stable training.
    \item Iso-Dreamer \cite{pan2022isodreamisolatingleveragingnoncontrollable}: decomposes scenes into action controllable and non-controllable components.
    \item DQN \cite{mnih2013playingatarideepreinforcement}, PPO \cite{schulman2017proximalpolicyoptimizationalgorithms}: Standard model-free algorithms.
\end{enumerate}

\subsection{\textbf{Metrics}}

\textit{Success Rate (SR \%)} as the percentage of runs where the ego-agent is able to achieve at least 90\% \textit{route completion} (RC \%) with zero instances of \textit{Infractions} which include the instances of going out-of-lane bounds and collisions with other traffic vehicles.

\textit{Episodic Return} is defined as the cumulative sum of the reward function (eq. \ref{reward_function_equation}) over a episode. $R_{0} = \sum_{t=0}^{T}r_{t}$

\subsection{\textbf{Results}}
\subsubsection{\textbf{Driving Scores and Episodic Returns}}
Table \ref{table:Driving-metrics} compares RL experts on standard driving metrics across all scenarios. Our model, \notcool, outperforms all baselines with a 20\% and 14.6\% improvement in \textit{SR} and \textit{RC}, respectively, while reducing \textit{Infraction/km} by 50\%. This demonstrates the effectiveness of \textit{DPA} and \textit{DiffDreamer}.

World model-based methods consistently surpass model-free approaches by explicitly learning an environment model, enabling more efficient planning. Among them, \textit{Dreamer-V3} outperforms \textit{Iso-Dreamer}, benefiting from enhanced robustness and normalization techniques over \textit{Dreamer-V2} \cite{hafner2022masteringataridiscreteworld}. PPO surpasses DQN by optimizing policies with stable updates and improving exploration through stochastic policies.

\begin{figure}[h]
    \centering
    \includegraphics[width=0.9\linewidth]{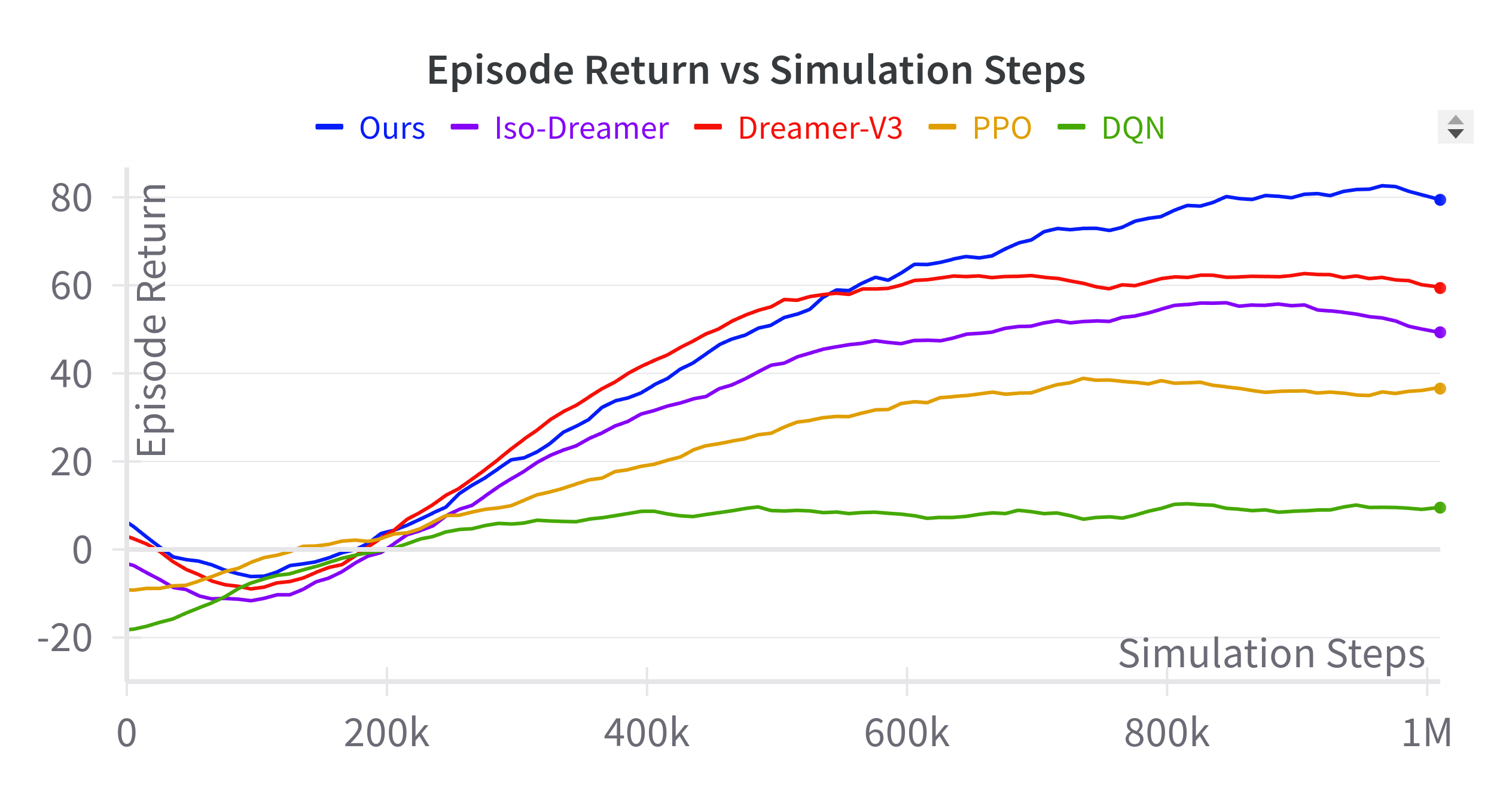}
    \caption{Episodic return of \notcool (Ours) and other different RL agents averaged over the scenarios. A moving average with a window size of 50 is applied}
    \label{fig:RewardsVsMethods}
\end{figure}

Fig. \ref{fig:RewardsVsMethods} further supports these findings, showing that \notcool achieves higher episodic returns over 1M simulation steps across all scenarios.
\label{driving-scores-&-episodic-returns}

\vspace{-5px}
\begin{table}[h] 
\caption{\textbf{Driving Metrics Comparison:} We compare the \notcool and the baselines on standard driving metrics across all the scenarios. We run each model on all scenarios with 3 random seeds.}
\scalebox{0.88}{ 
\begin{tabular}{|p{2.7cm}|p{1.3cm}|p{2.5cm}|p{1.35cm}|}  
\hline
\textbf{Model}  & \textbf{SR(\%) $\uparrow$}& \textbf{Infraction/Km $\downarrow$}& \textbf{RC(\%) $\uparrow$} \\
\hline
DQN \cite{mnih2013playingatarideepreinforcement}  & 0.0 & 6.27  & 27.63\\
\hline
PPO \cite{schulman2017proximalpolicyoptimizationalgorithms}  & 16.66 & 4.61 & 50.02 \\
\hline
Iso-Dreamer \cite{pan2022isodreamisolatingleveragingnoncontrollable}  & 56.66 & 1.65 & 60.33\\
\hline
Dreamer-V3 \cite{hafner2024masteringdiversedomainsworld}  & 63.33 & 1.52  & 67.53\\
\hline
\notcool (ours)  & \textbf{83.33} & \textbf{0.70}  & \textbf{82.13}\\
\hline
\end{tabular}
}
\label{table:Driving-metrics}
\end{table}

\begin{table}[h] 
\caption{\textbf{Driving Metrics Comparison for Ablations:} We compare the \notcool and the baselines, along with combining (DPA) with other world-model based methods on standard driving metrics across all the scenarios. We run each model on all scenarios with 3 random seeds.}
\scalebox{0.88}{ 
\begin{tabular}{|p{2.7cm}|p{1.3cm}|p{2.5cm}|p{1.35cm}|}  
\hline
\textbf{Model}  & \textbf{SR(\%) $\uparrow$}& \textbf{Infraction/Km $\downarrow$}& \textbf{RC(\%) $\uparrow$} \\
\hline
Iso-Dreamer \cite{pan2022isodreamisolatingleveragingnoncontrollable}  & 56.66 & 1.65  & 60.33\\
\hline
Dreamer-V3 \cite{hafner2024masteringdiversedomainsworld}  & 63.33 & 1.52 & 67.53 \\
\hline
\textit{DPA} + Iso-Dreamer  & 73.33 & 1.17 & 70.87\\
\hline
\textit{DPA} + Dreamer-V3  & 83.33 & 0.97  & 75.09\\
\hline
\notcool (ours)  & \textbf{83.33} & \textbf{0.70}  & \textbf{82.13}\\
\hline
\end{tabular}
}
\label{table:Ablation-Driving-metrics}
\end{table}
\vspace{10px}
\subsubsection{\textbf{Prediction Fidelity}}
Table \ref{table:FIDandFVD} shows a quantitative comparison of world models prediction fidelity. \notcool outperforms in temporal consistency and fidelity metrics by 72 and 290 in FID and FVD scores respectively. Unlike auto-regressive models, which degrade over time, SVD models enhance temporal consistency by incorporating temporal attention.
Fig. \ref{fig:action-controllability} qualitatively highlights the high fidelity prediction of our world model conditioned on an input trajectory. 
\label{prediction-fidelity-results}


\begin{table}[h] 
\caption{Comparison of prediction fidelity scores of different world models over CARLA driving sequences}
\scalebox{0.92}{ 
\begin{tabular}{|p{2.5cm}|p{2cm}|p{2cm}|}  
\hline
\textbf{World Model}  & \textbf{FID $\downarrow$}& \textbf{FVD $\downarrow$}\\
\hline
DriveGAN \cite{kim2021drivegancontrollablehighqualityneural}  & 67.1 & 281.9\\
\hline
Iso-Dreamer \cite{pan2022isodreamisolatingleveragingnoncontrollable}  & 102.24 & 421.56\\
\hline
Dreamer-V3 \cite{hafner2024masteringdiversedomainsworld}  & 89.29 & 324.07\\
\hline
\textit{DiffDreamer}  & \textbf{17.09} & \textbf{130.39}\\
\hline
\end{tabular}
}
\label{table:FIDandFVD}
\end{table}

\vspace{-0px}
\section{ABLATION STUDIES} \label{ablations}
\subsection{\textbf{DPA analysis}}
Table \ref{table:Ablation-Driving-metrics} evaluates \textit{DPA} within the \textit{Iso-Dreamer} and \textit{Dreamer-V3} frameworks by comparing performance with their SAC actor versus when replaced with \textit{DPA} \underline{(rows 1 vs. 3)} and \underline{(rows 2 vs. 4)}. Models using \textit{DPA} outperform by at least 10\% and 8\% on \textit{SR} and \textit{RC} metrics, highlighting its impact on trajectory prediction. Furthermore, Fig. \ref{multi-modal-result} illustrates \textit{DPA}'s ability to model complex, multi-modal action distributions, demonstrating its adaptability and broader potential in reinforcement learning and decision-making tasks.
\label{diff-policy-ablation}
\subsection{\textbf{\textit{DiffDreamer} Analysis}}
Table \ref{table:Ablation-Driving-metrics} evaluates prediction fidelity and environmental dynamics across different world models using a uniform \textit{DPA} policy. This isolates the impact of each model, allowing a focused assessment of its ability to capture environmental dynamics and influence trajectory prediction. Results in \underline{(rows 2, 3, and 4)} highlight \textit{DiffDreamer}'s superior performance, achieving a 20\% and 14.6\% improvement in \textit{SR} and \textit{RC}, respectively. This underscores the importance of accurate future state predictions and validates the use of SVD as the foundation of our world model framework.
\label{wm-ablation}
\subsection{\textbf{Context Length analysis}}
\begin{figure}[H]
    \centering
    \includegraphics[width=0.9\linewidth]{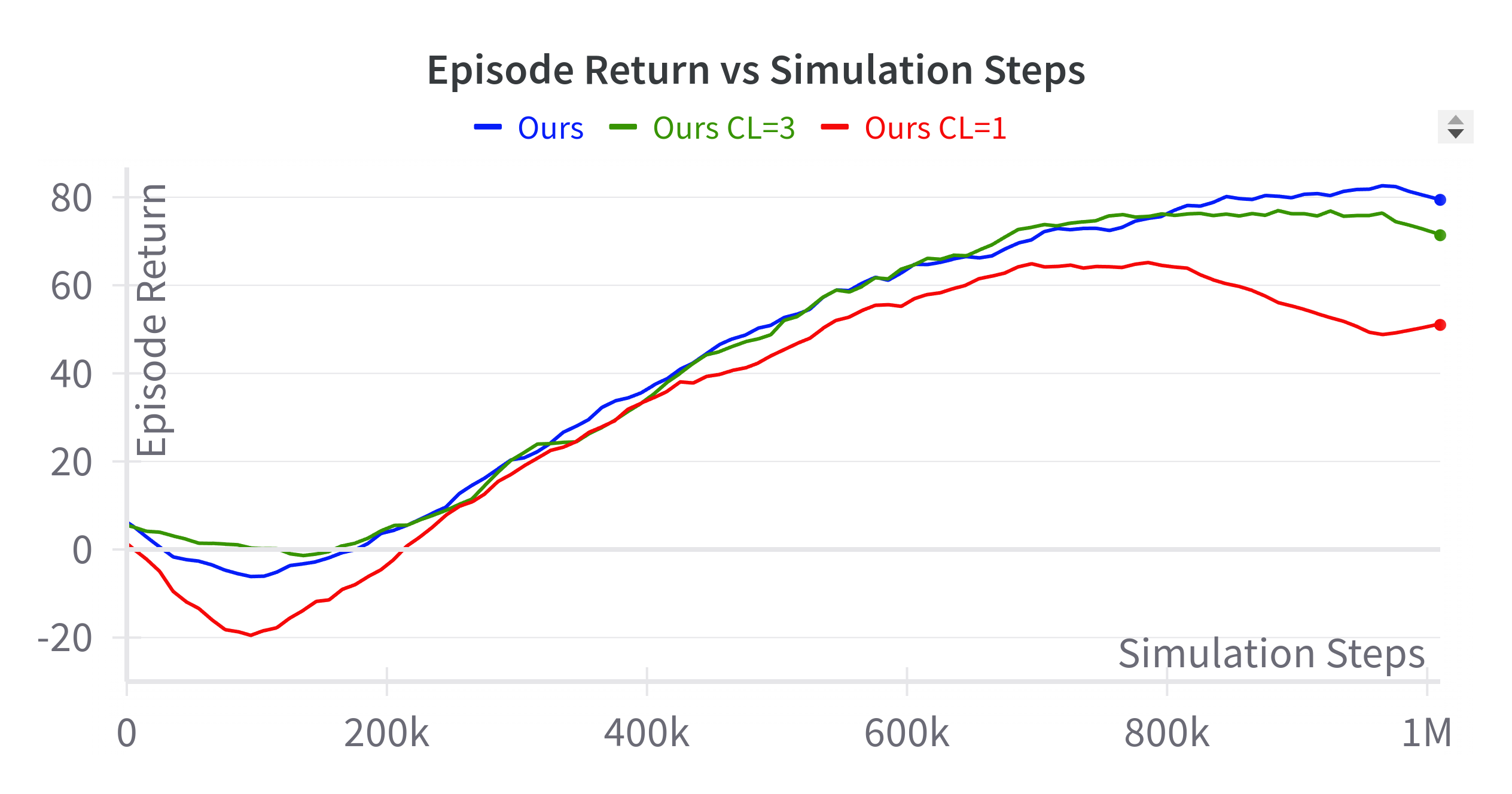}
    \caption{Episodic return of \notcool (Ours with \textit{CL}=5) with different \textit{CL} averaged over the scenarios. A moving average with a window size of 50 is applied. \textit{Context Length} refers to the number of past frames used for state encoding.}
    \label{fig:RewardsVsCL}
\end{figure}

To evaluate the impact of POMDP approximation using the \textit{State Encoder}, Fig. \ref{fig:RewardsVsCL} compares episodic returns for \notcool across different \textit{context lengths (CL)}, with $P = 5$ fixed for \textit{DiffDreamer}. The performance declines at $CL = 1$ due to insufficient historical information, limiting state inference in partially observable environments. In contrast, longer context lengths $(CL = 3, 5)$ capture richer temporal information, improving state estimation and decision-making, thereby enhancing performance.

\section{CONCLUSION \& FUTURE WORK} \label{Conclusion}
\textbf{Conclusion:} In this work, we introduced \notcool, a world model-based reinforcement learning framework that combines a high-fidelity world model (\textit{DiffDreamer}) with a multi-modal diffusion policy actor (\textit{DPA}) for long-horizon trajectory generation in autonomous driving. \textit{DiffDreamer} enables accurate future prediction in image state space, improving sample efficiency and reducing reliance on online interactions. Through iterative training, our approach captures diverse behavioral modes while enhancing learning stability. Experimental results highlight its superiority over single-Gaussian policies and other world models, demonstrating its ability to generate diverse trajectories and predict accurate future observations.

\textbf{Future Work:} We aim to extend \notcool to other domains such as robotic manipulation, where accurate long-horizon planning and multimodal prediction are crucial. Expanding the versatility of our approach across domains will further validate its generalizability and adaptability.

\printbibliography

\end{document}